\documentclass[final]{cvpr}

\usepackage{times}
\usepackage{epsfig}
\usepackage{graphicx}
\usepackage{amsmath}
\usepackage{amssymb}

\usepackage{multirow}
\usepackage{subfigure}
\usepackage[english]{babel}
\usepackage{threeparttable}

\usepackage[pagebackref=false,breaklinks=true,letterpaper=true,colorlinks,bookmarks=false]{hyperref}

\def\cvprPaperID{6677} 
\def\confYear{CVPR 2021}

\begin{document}
	
	\title{PISE: Person Image Synthesis and Editing with Decoupled GAN}
	\author{\parbox{\textwidth}{\centering 
			Jinsong Zhang$^{1}$,
			Kun Li$^{1}$\thanks{Corresponding author},
			Yu-Kun Lai$^{2}$, 
			Jingyu Yang$^{1}$
		}
		\\
		{\parbox{\textwidth}{\centering $^1$Tianjin University, China	$^2$ Cardiff University, United Kingdom
			}
		}
	}
	
	\maketitle
	\pagestyle{empty}
	\thispagestyle{empty}

	\begin{abstract}
		Person image synthesis, e.g., pose transfer, is a challenging problem due to large variation and occlusion. Existing methods have difficulties predicting reasonable invisible regions and fail to decouple the shape and style of clothing, which limits their applications on person image editing. In this paper, we propose PISE, a novel two-stage generative model for Person Image Synthesis and Editing, which is able to generate realistic person images
		with desired poses, textures, or semantic layouts. For human pose transfer, we first synthesize a human parsing map aligned with the target pose to represent the shape of clothing by a parsing generator, and then generate the final image by an image generator. To decouple the shape and style of clothing, we propose joint global and local per-region encoding and normalization to predict the reasonable style of clothing for invisible regions. We also propose spatial-aware normalization to retain the spatial context relationship in the source image. The results of qualitative and quantitative experiments demonstrate the superiority of
		our model on human pose transfer. Besides, the results of texture transfer and region
		editing show that our model can be applied to person image editing. The code is available for research purposes at \href{https://github.com/Zhangjinso/PISE}{https://github.com/Zhangjinso/PISE}.
	\end{abstract}
	
	
	\vspace{-0.5cm}
	\section{Introduction}
	
	Person image synthesis is a challenging problem in computer vision and computer graphics, which has great application potentials in image editing, video generation, virtual try-on, \emph{etc}. Human pose transfer \cite{adgan,gfla,bi,xinggan,patn}, \emph{i.e.}, synthesizing a new image for the same person in a target pose, is an active topic in person image synthesis. 
	
	\begin{figure}[!ht]
		
		\centering
		\includegraphics[scale=0.43]{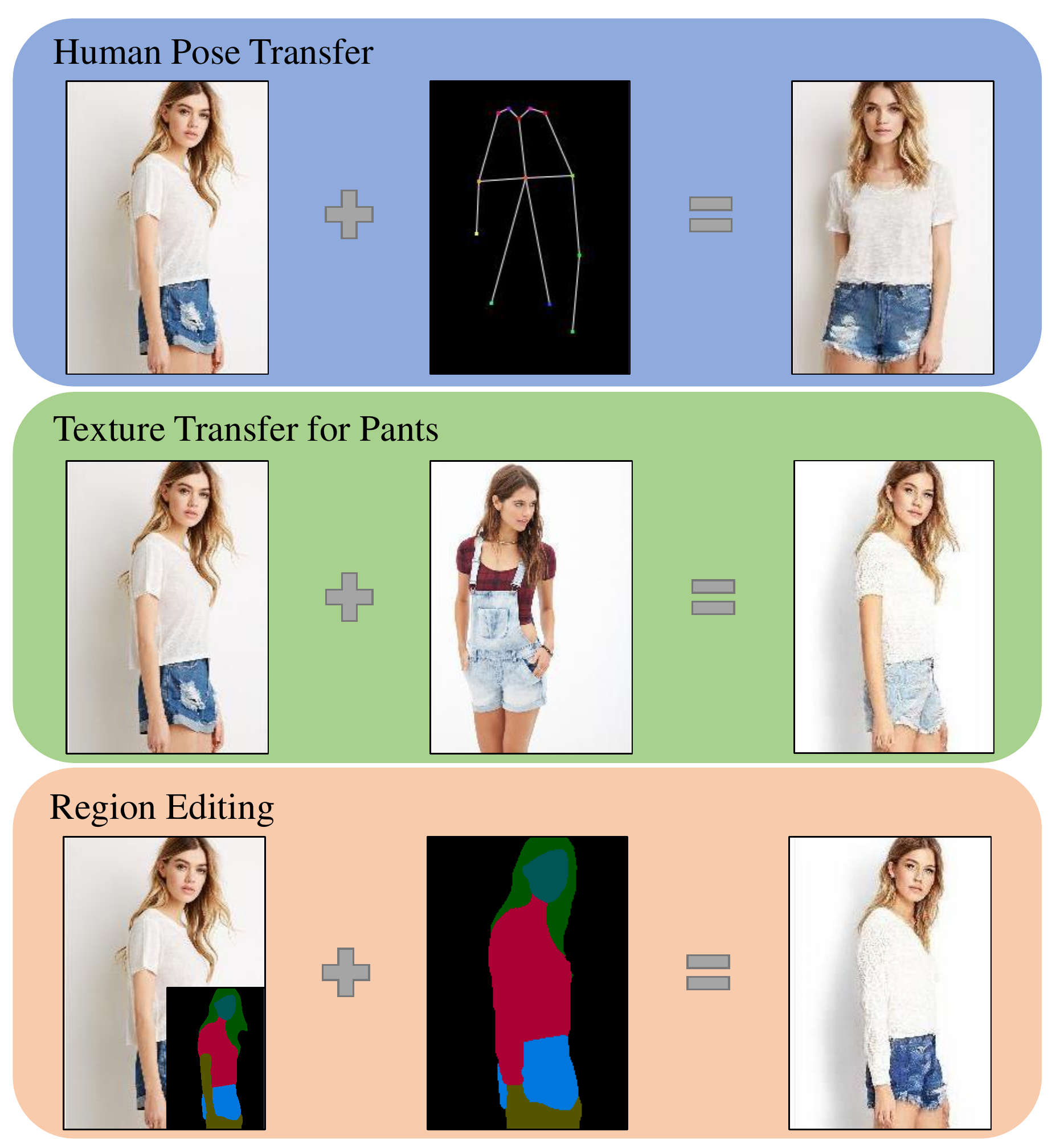}
		\caption{Our model PISE allows to transfer new pose or texture to a single person image, and also enables region editing. }
		\label{fig_pres}
	\end{figure}
	
	
	Recently, Generative Adversarial Networks (GANs)~\cite{gan} achieve great success in human pose transfer.
	Many methods directly learn the mapping from the source image and pose to the target image using neural networks \cite{patn, bi, xinggan,pona}. Most of these methods utilize a two-branch (pose branch and image branch) framework to transfer the feature of the source image from the source pose to the target pose. However, by taking keypoints as the pose representation, it is difficult to predict a sharp and reasonable image with sparse correspondences when the source pose and the target pose have large differences. 
	To deal with this problem, flow-based methods \cite{flow19, gfla} estimate an appearance flow to obtain denser correspondences, which is used to warp the source image or its feature to align with the target pose. 
	The final image delivered by refining the warped image or decoding the warped image feature is a rearrangement of the source image elements.
	Thus, the generated image can preserve details of the source image, but the invisible region due to occlusion is not satisfactorily recovered.
	To predict invisible regions, some methods \cite{adgan, pinet} introduce human parsing maps to human pose transfer. 
	The human parsing map provides semantic correspondence to synthesize the final image and enables  applications of person image editing. However, these methods cannot disentangle the shape and style information (\emph{e.g.}, the category and texture of clothing) and fail to preserve spatial context relationships. For flexible and detailed editing, it is better to disentangle the shape and style. Meanwhile, preserving spatial information of the source image and reasonable prediction of invisible regions are also important for producing the desired output for human pose transfer.
	
	The aforementioned methods encounter three challenges to synthesize satisfactory images: 1) the coupling of the shape and style of clothing, 2) potential uncertainties in invisible regions, and 3) loss of spatial context relationships.
	
	To address these problems, we propose a novel Decoupled GAN for person image synthesis and editing.
	Instead of directly learning a mapping from the source to the target,
	we take the human parsing map as the intermediate result to provide semantic guidance to predict a reasonable shape of clothing. 
	We propose joint global and local per-region encoding and normalization to control the texture style on a semantic region basis, better utilizing information for both visible and invisible regions.
	Specifically, for the region visible in the source image, we use the local feature of the corresponding region to predict the style of clothing. 
	For the region invisible in the source image but visible in the target image, we obtain the global feature of the source image to predict the reasonable style of clothing, which can well deal with the generation of invisible regions. 
	Benefiting from the human parsing map and per-region texture control, the shape and the style of clothing are disentangled for more flexible editing. 
	Besides, to preserve the spatial context relationship, we propose a novel spatial-aware normalization to transfer spatial information of the source image to the generated image.
	After per-region normalization and spatial-aware normalization, the generated target feature passes through a decoder to output the final image. 
	Figure \ref{fig_pres} shows some applications of our model. 
	
	The main contributions of this work are summarized as follows:
	\begin{itemize} 
		\item We propose a two-stage model with per-region control to decouple the shape and style of clothing. Experimental results on human pose transfer, texture transfer, and region editing show the flexibility and superior performance of our person image synthesis and editing method.
		\item We propose joint global and local per-region encoding and normalization  to predict the reasonable style of clothing for invisible regions, and preserve the original style of clothing in the target image.
		\item We propose a spatial-aware normalization  to retain the spatial context relationship in the source image, and transfer it by modulating the scale and bias of the generated image feature.
	\end{itemize}

	
	\section{Related Work}

	\subsection{Image Synthesis}
	
	In recent years, Generative Adversarial Networks (GANs) \cite{gan} have made great success in image synthesis \cite{deepfillv1, deepfillv2, structureflow}. 
	Isola \emph{et al.} \cite{pix2pix} first introduced conditional GANs \cite{cgan} to solve the image-to-image generation task, which was extended to high-resolution image synthesis \cite{pix2pixhd}. Zhu \emph{et al.} proposed an unsupervised method with cycle consistency to transfer images between two domains without paired data. 
	StyleGAN \cite{stylegan} used adaptive instance normalization (AdaIN) \cite{adain} to achieve scale-specific control of image synthesis.
	SPADE \cite{spade} adopted spatially-adaptive normalization to synthesize new images given semantic input by modulating the activations in normalization layers.
	Zhu \emph{et al.} \cite{smis}  leveraged group convolution and designed a Group Decreasing Network to alleviate memory access cost problem.
	SEAN \cite{sean} improved SPADE by proposing per-region encoding to control the style of individual regions in the generated images.      
	However, these methods have limited editable capacity in human pose transfer due to sparse correspondence of keypoints and large variation in pose and texture. 
	In this paper, we propose a two-stage model to obtain semantic guidance to achieve more controllable image synthesis. 
	
	\begin{figure*}[!ht]
		
		\centering
		\includegraphics[height=70mm,width=150mm]{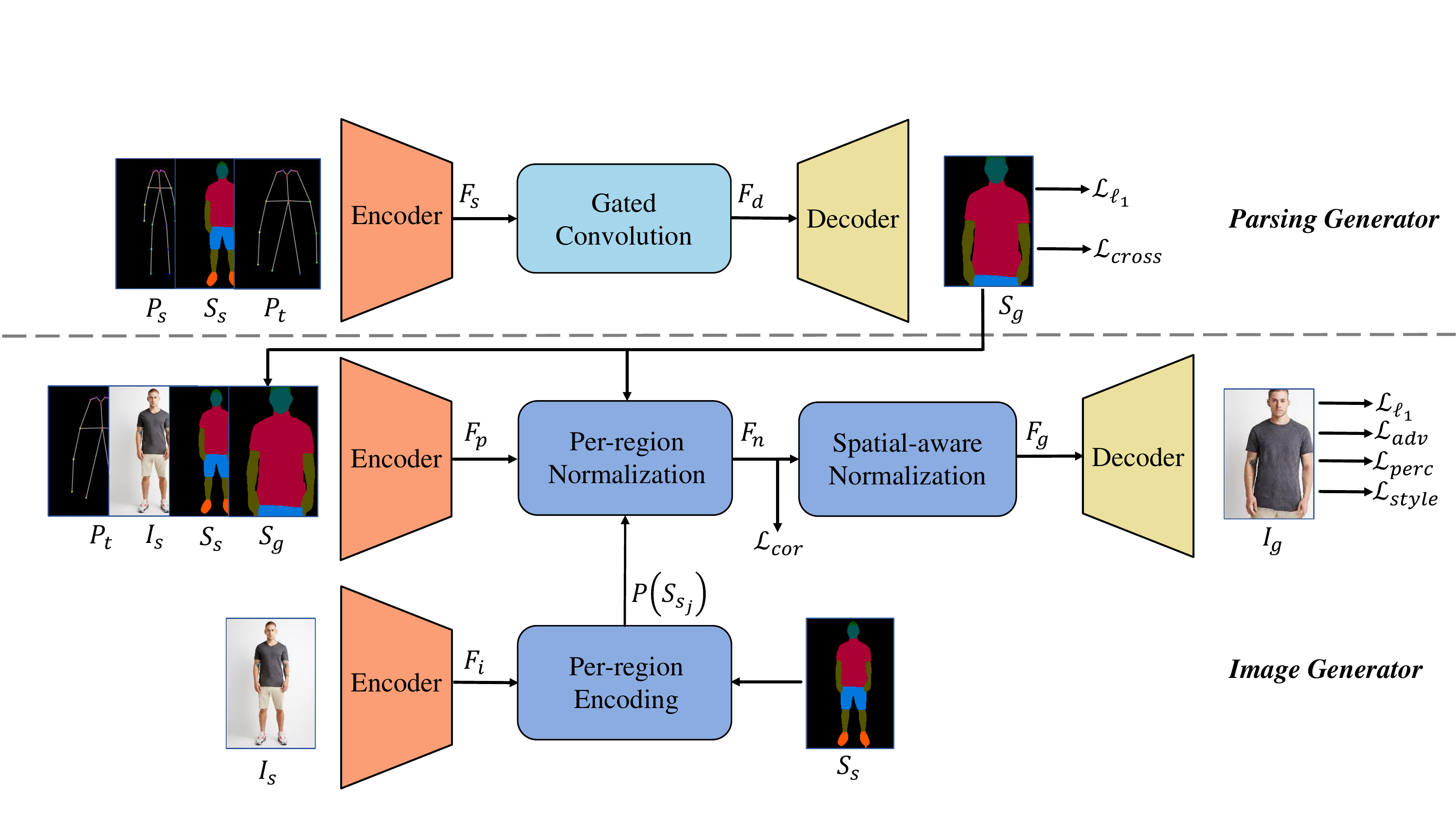}
		\caption{Overview of our model. }
		\label{fig_net}
		\vspace{-0.2cm}
	\end{figure*}
	
	\subsection{Human Pose Transfer}
	
	Human pose transfer is a highly active topic in computer vision and computer graphics. 
	PG$^2$ \cite{pg2} firstly introduced this problem and utilized a coarse-to-fine framework to alleviate the challenging generation problem. 
	It concatenated 
	the source image, the source pose, and the target pose as inputs to learn the target image, which leads to feature misalignment.
	Some methods used a two-branch framework with the image branch and pose branch to deal with the misalignment between the source and target images.
	Zhu \emph{et al.} \cite{patn} proposed to transfer image information from the source pose to the target pose progressively with a local attention mechanism.
	Tang \emph{et al.} \cite{xinggan} modeled appearance information and shape information with two novel blocks.
	Li \emph{et al.} \cite{pona} designed pre-posed image-guided pose feature update and post-posed pose guided image feature update to better utilize the pose and image features.
	These methods used keypoints as their pose representation and focused on transferring image information with the guidance of pose information.
	Therefore, only sparse correspondence between the source image and the target image can be obtained, which is difficult to transfer image information from the source pose to the target pose.
	To provide semantic correspondence between the source image and the target image, some methods  introduced the human parsing map as semantic guidance.
	Dong \emph{et al.} \cite{softgated} used a two-stage model, which first synthesized a target semantic segmentation map and then rendered textures from the original image using a soft-gated warping block.
	Han \emph{et al.} \cite{clothflow} first synthesized a target human parsing map, then estimated a dense flow to warp source image feature and finally refined the image using a U-Net-based \cite{unet} network.
	These flow-based methods can preserve some details, but cannot cope well with large occlusion, which limits their application on person image synthesis.
	Men \emph{et al.} \cite{adgan} used human parsing maps for attribute-controllable person image synthesis.
	Zhang \emph{et al.} \cite{pinet} introduced gated convolution to learn a dynamic feature selection mechanism and adaptively deform the image layer by layer.
	However, these methods fail to disentangle the shape and style information and cannot edit the image flexibly. 
	Our model generates more reasonable and realistic person images  with the consistencies of both shape and style. Besides, our model can edit the image more flexibly by disentangling the shape and style information.

	\vspace{-0.2cm}	
	\section{Method}
	
	
	As shown in Figure \ref{fig_net}, our approach consists of two generators: a parsing generator and an image generator. The inputs of the parsing generator are the source pose $P_s$, the target pose $P_t$, and the source parsing map $S_s$.
	The parsing generator estimates a human parsing map $S_g$ aligned with the target pose $P_t$. This allows the image editing of the shape of the final image to be controllable. 
	The pose representation includes 18 human keypoints extracted by Human Pose Estimator (HPE) \cite{hpe}, which has 18 channels and encodes the locations of 18 joints of a human body. The source parsing map $S_s$ is extracted by Part Grouping Network (PGN) \cite{pgn} from the source image $I_s$. 
	To clean incorrect labels (\emph{e.g.}, left leg and right leg) and reduce the number of categories, we re-organize the map from 21 categories to 8 categories: hair, upper clothes, dress, pants, face, upper skin, leg, and background.
	The image generator synthesizes a high-quality image $I_g$ of the reposed person conditioned on the human parsing map $S_g$. The inputs of the image generator are the source image $I_s$, the source parsing map $S_s$, the generated parsing map $S_g$ and the target pose $P_t$. 
	To provide detailed control of the styles in individual regions, we propose joint global and local per-region encoding and normalization to decouple the style and shape. With the generated parsing map $S_g$ and per-region style control, we decouple the shape and style of clothing to facilitate image editing tasks.
	Besides, we propose spatial-aware normalization to retain the spatial context relationship. Note that the model can deal with other tasks, \eg, texture transfer and region editing.
	
	\vspace{-0.2cm}
	\subsection{Parsing Generator}
	\vspace{-0.1cm}
	
	The parsing generator is responsible for generating the human parsing map aligned with the target pose while keeping the clothing style and body shape of the person in the source image. 
	The inputs, the source pose $P_s$, the target pose $P_t$, and the source parsing $S_s$, are first embedded into a  latent space by an encoder, which consists of $M$ down-sampling convolutional layers ($M$ = 4 in our case). 
	Inspired by PINet \cite{pinet}, instead of applying residual blocks \cite{resnet} like previous methods \cite{softgated, clothflow}, we use gated convolution to deform the feature $F_s$ of source human parsing map $I_s$ from the source pose to the target pose to avoid the drawback of vanilla convolution that treats all the pixels as valid information. 
	The formulation of gated convolution is 
	\vspace{-0.2cm}
	\begin{equation}
	\begin{split}
	O_{x,y} = \phi(\sum_{i=-k_h}^{k_h}\sum_{j=-k_w}^{k_w} u_{k_h+i,k_w+j} \cdot I_{y+i,x+j}) \cdot \\ \sigma(\sum_{i=-k_h}^{k_h}\sum_{j=-k_w}^{k_w} v_{k_h+i,k_w+j} \cdot I_{y+i,x+j}),
	\end{split}
	\end{equation}
	where $\cdot$ denotes the element-wise multiplication of two feature maps. $I_{x,y}$ and $O_{x,y}$ are the input and output at position $(x, y)$,
	$k_h$ = $(k_{sh}-1)/2$ and $k_w$ = $(k_{sw}-1)/2$. $k_{sh}$ and $k_{sw}$ are the kernel sizes (\emph{e.g.}, 3 $\times$ 3).  $\sigma$ denotes sigmoid function which ensures the output gating values are between 0 and 1. 
	$\phi$ denotes the activation function. $u$ and $v$ are two different convolutional filters. 
	
	Gated convolution can learn a dynamic selection mechanism for each spatial location, which is suitable for unaligned generation tasks \cite{deepfillv1, deepfillv2, pinet}.
	Finally, with the deformed parsing feature $F_d$, the generated human parsing map $S_g$ is delivered by a decoder with standard configuration. 
	The detailed network design can be found in the supplementary material.
	
	\vspace{-0.2cm}
	\subsection{Image Generator}
	\vspace{-0.1cm}
	
	The image generator aims at transferring the textures of individual regions in the source image to the generated parsing map. From another point of view, this can be seen as a semantic map-to-image translation problem conditioned on 
	the source image.
	Inspired by SEAN \cite{sean}, 
	we first extract per-region styles of the source image $I_s$ with the source human parsing map $P_s$, which are then transferred using normalization techniques.
	Because some visible regions in the target image are invisible in the source image due to large variation in pose, the number of regions in our generated parsing map is different from that of the source image. SEAN sets the styles of invisible regions to zero, which ignores the implied relationship between visible and invisible regions of person images (\emph{e.g.}, a man in a coat is more likely to wear trousers than shorts). 
	Therefore, instead of using only local region-wise average pooling, we propose \emph{joint global and local per-region average pooling} to extract the style of the region in the source image.
	We then concatenate the source image $I_s$, the source parsing map $S_s$, the generated parsing map $S_g$ and the target pose $P_t$ in depth (channel) dimension and extract its feature $F_p$.
	Finally, similar to existing normalization techniques, we control the per-region style of $F_p$ by modulating its scale and bias.
	However, previous normalization techniques lose the spatial information of the source image. To solve this problem, we propose a \emph{spatial-aware normalization} method to preserve the spatial context relationship of the source image. 
	After spatial-aware normalization, the desired person image $I_g$ is delivered by a decoder.
	
	\vspace{-0.2cm}
	\subsubsection{Per-region encoding}
	\vspace{-0.1cm}
	\label{per-region encoding}
	
	Given the source image $I_s$, we first extract its feature map  using a bottleneck structure with 4 down-sampling convolution layers and 2 transposed convolution layers. 
	The output of encoder $F_i$ 
	with 256 channels contains the spatial context relationship and style of the source image.
	Intuitively,
	the styles of individual regions are independent of their shapes. With the source parsing map $S_s$, we disentangle the shape and style by extracting style information using average pooling to remove the shape information. 
	To preserve the style information visible in the source image and predict a reasonable style for invisible regions conditioned on the source image, we propose \emph{joint global and local per-region average pooling} to extract the style of the source image, which is formulated as
	\begin{equation}
	P(S_{s_j}) = \begin{cases}
	\mathop{avg}\limits_{w,h} (F_i \cdot S_{s_j}), & \sum S_{s_j} > 0  \\
	\mathop{avg}\limits_{w,h} (F_i),& \sum S_{s_j} \leq 0 \\
	\end{cases},
	\end{equation}
	where $avg$($\cdot$) denotes average pooling in the spatial dimension. $w$ and $h$ are the width and height of the feature map $F_i$. $j$ indicates the category index, and $S_{s_j}$ is the source semantic map w.r.t. category $j$.
	The first case considers local average pooling where category $j$ appears in the source image, and the second case is global average pooling for unseen categories.
	The size of output $P(S_{s_j})$ after average pooling is 256 $\times$ $N$, where $N$ is the number of label categories (8 in our case). 
	
	\vspace{-0.2cm}
	\subsubsection{Per-region normalization}
	\vspace{-0.1cm}
	
\begin{figure}[!t]
		
		\centering
		\includegraphics[scale=0.28]{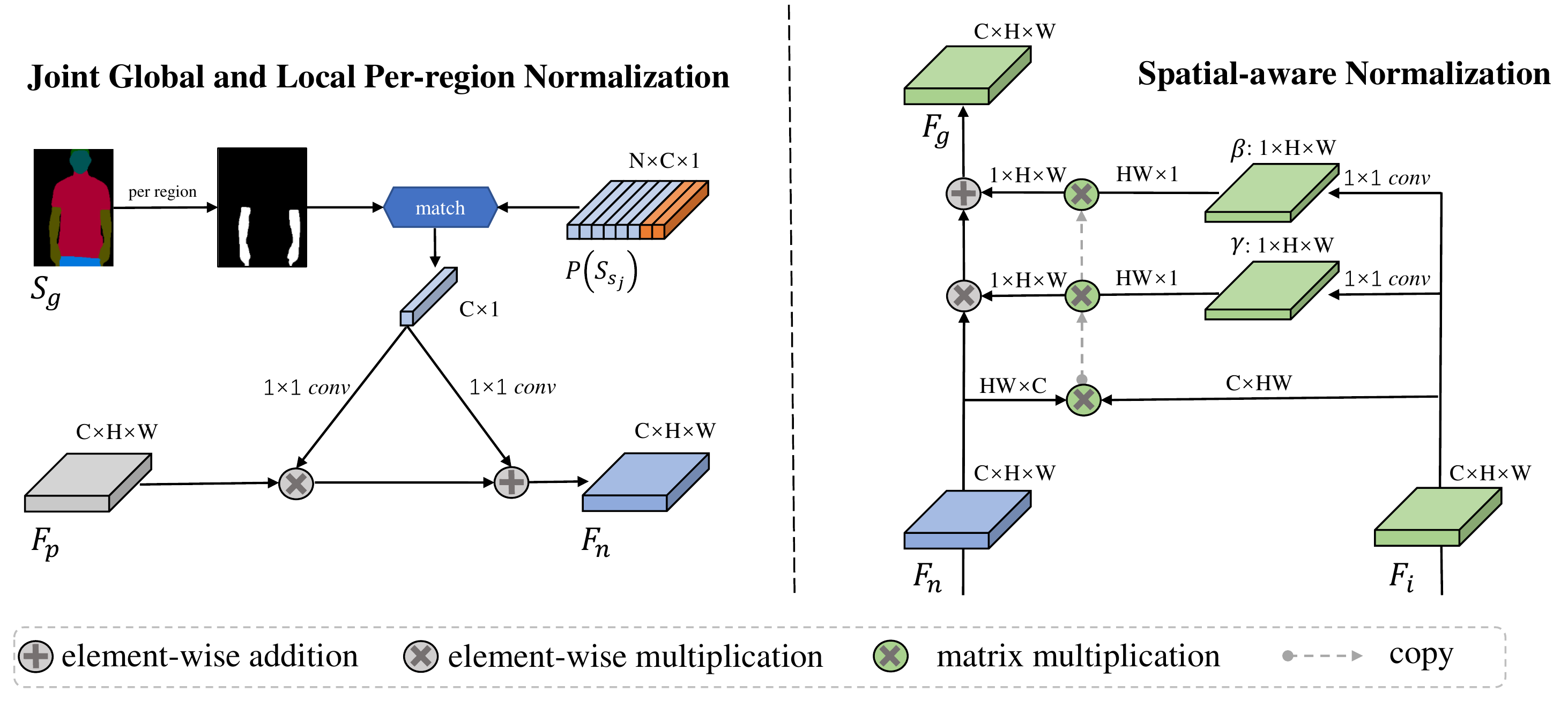}
		\caption{Details of per-region normalization and spatial-aware normalization in our model. }
		\label{fig_sn}
		\vspace{-0.2cm}
	\end{figure}

	The newly generated feature $F_p$ from a basic encoder contains the information of the inputs and aligns with the generated parsing map $S_g$.
	With the per-region style code $P(S_{s_j})$ w.r.t. label category $j$, we can transfer the style to the newly generated feature $F_p$ by modulating its scale and bias. 
	As shown in Figure \ref{fig_sn}, for each region $S_{g_j}$ in the generated parsing map $S_g$, we find the corresponding region style in $P(S_{s_j})$. 
	Thanks to our joint global and local per-region encoding, the style codes of all the regions can be found in $P(S_{s_j})$.
	Then we use two fully connected layers to predict the scale and bias for $F_p$, which are then applied for per-region normalization.
	The generated feature $F_p$ is updated to $F_n$ that contains the per-region styles of the source image.

	\vspace{-0.2cm}
	\subsubsection{Spatial-aware normalization}
	\label{sec_sn}
	\vspace{-0.1cm}
	
	As illustrated in Section \ref{per-region encoding}, the style included in $P(S_{s_j})$ loses the spatial context relationship of the source image due to average pooling. It is difficult to make the generated images preserve the details (especially spatial details) in the source image.
	As shown in Figure \ref{fig_sn}, to preserve the spatial context relationship in the source image, in addition to extracting styles by average pooling in the spatial dimension, we also  extract spatial scale ($\gamma$) and bias ($\beta$) from the source image code $F_i$ using 1 $\times$ 1 convolution layers.
	We retain spatial context relationships using $\gamma$ and $\beta$. However, due to the misalignment between the source image and the target image, how to transfer $\gamma$ and $\beta$ to the generated image is a challenging problem. 
	To tackle this problem, we try to compute the correspondence between the feature after per-region normalization $F_n$ and the feature of the source image $I_s$. 
	We first use a correspondence loss to constrain the similarity between $F_n$ and the pre-trained VGG-19 \cite{vgg} feature of the target image. 
	With the correspondence loss, $F_n$ can be more aligned with the feature of the target image in the latent space, which constrains them to be in the same domain. 
	Then, we compute the correspondence between $F_n$ and the VGG-19 feature of the source image $I_s$. 
	We use the correspondence layer \cite{he2018deep} to compute a correlation matrix
	\begin{equation}
	\mathcal{M}(p_1, p_2) = \frac{{F_n}(p_1)^T {\phi_i}(I_s)(p_2)}{||{F_n}(p_1)|| \ ||{\phi_i}(I_s)(p_2)||},
	\vspace{-0.1cm}
	\end{equation}
	where $\phi_i$ denotes the activation map of the $i$-th layer of the VGG-19 network.
	We use $conv3\_1$ in our experiments. ${F_n}(p_1)$ and ${F_n}(p_2)$ denote the channel-wise centralized features of $F_n$ at the positions $p_1$ and $p_2$, respectively.
	
	The $\gamma$ and $\beta$ that denote spatial context relationships at each position
	can be transformed from the source image to the target image by multiplying them with the correlation matrix $\mathcal{M}$.
	Then, $F_n$ is further updated to $F_g$ by modulating the scale and bias.
	Therefore, after passing through the spatial-aware normalization module, $F_g$ retains both the style  and spatial relationships of the source image. Finally, The decoder outputs the final image $I_g$.

	\begin{figure*}[!ht]
		
		\centering
		{
			\begin{minipage}[t]{0.5\linewidth}
				\centering
				\includegraphics[scale=0.25]{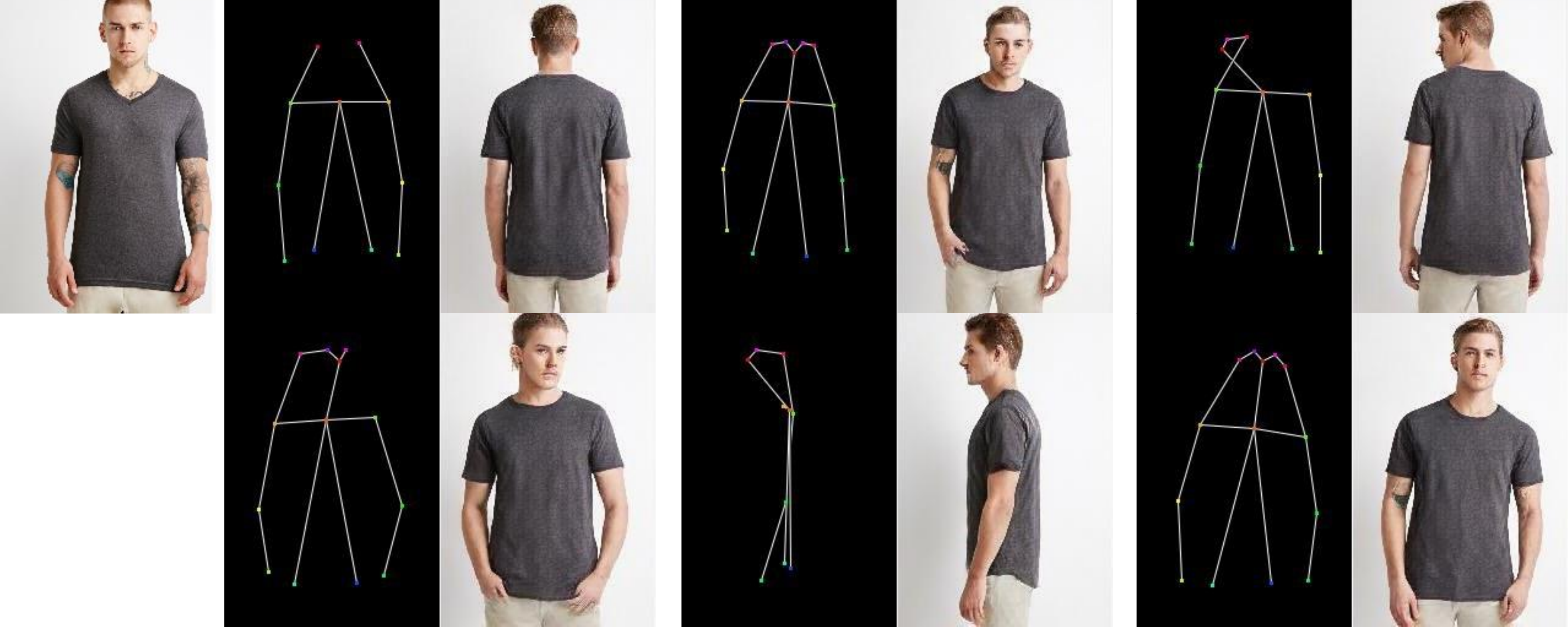}
			\end{minipage}%
			\begin{minipage}[t]{0.5\linewidth}
				\centering
				\includegraphics[scale=0.25]{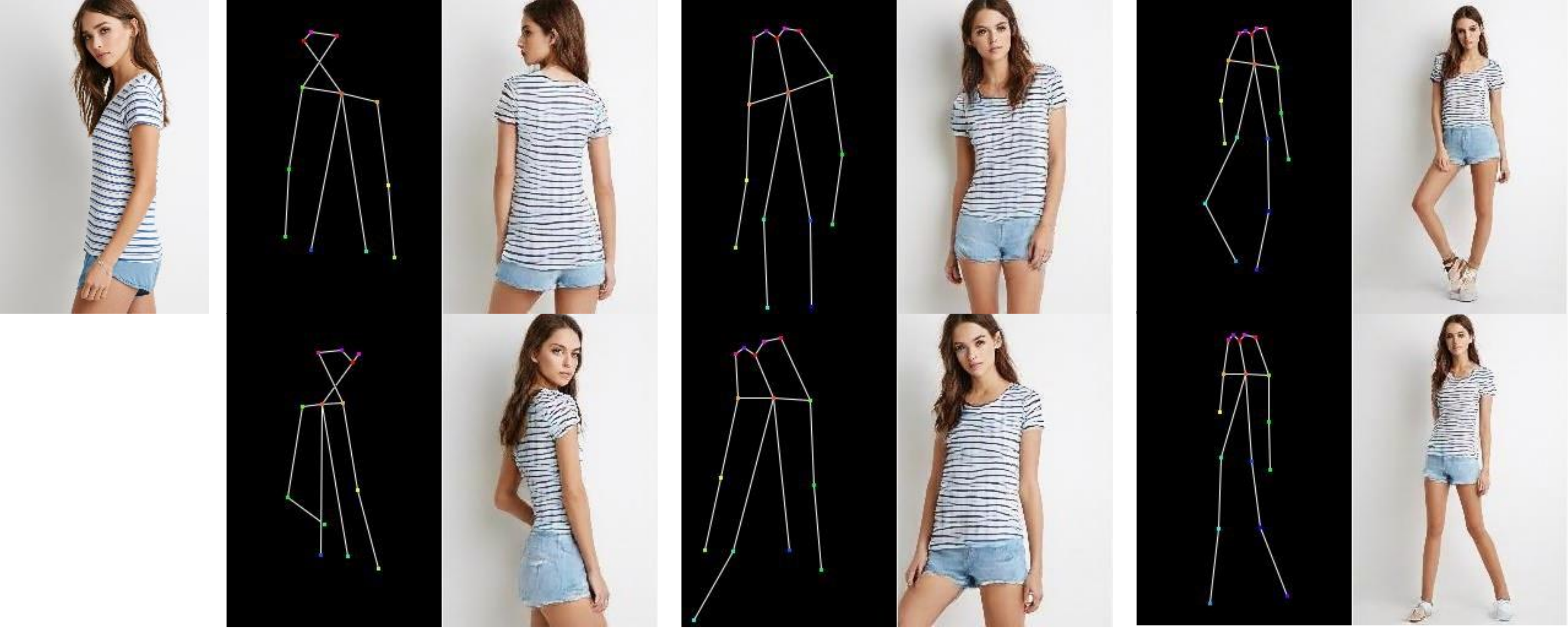}
			\end{minipage}%
		}%
		
		\caption{Our results of person image synthesis in arbitrary poses. }
		\label{fig_arb}
		\vspace{-0.2cm}
	\end{figure*}

	\begin{figure*}[!ht]
		
		\centering
		\begin{center}
			\includegraphics[scale=0.47]{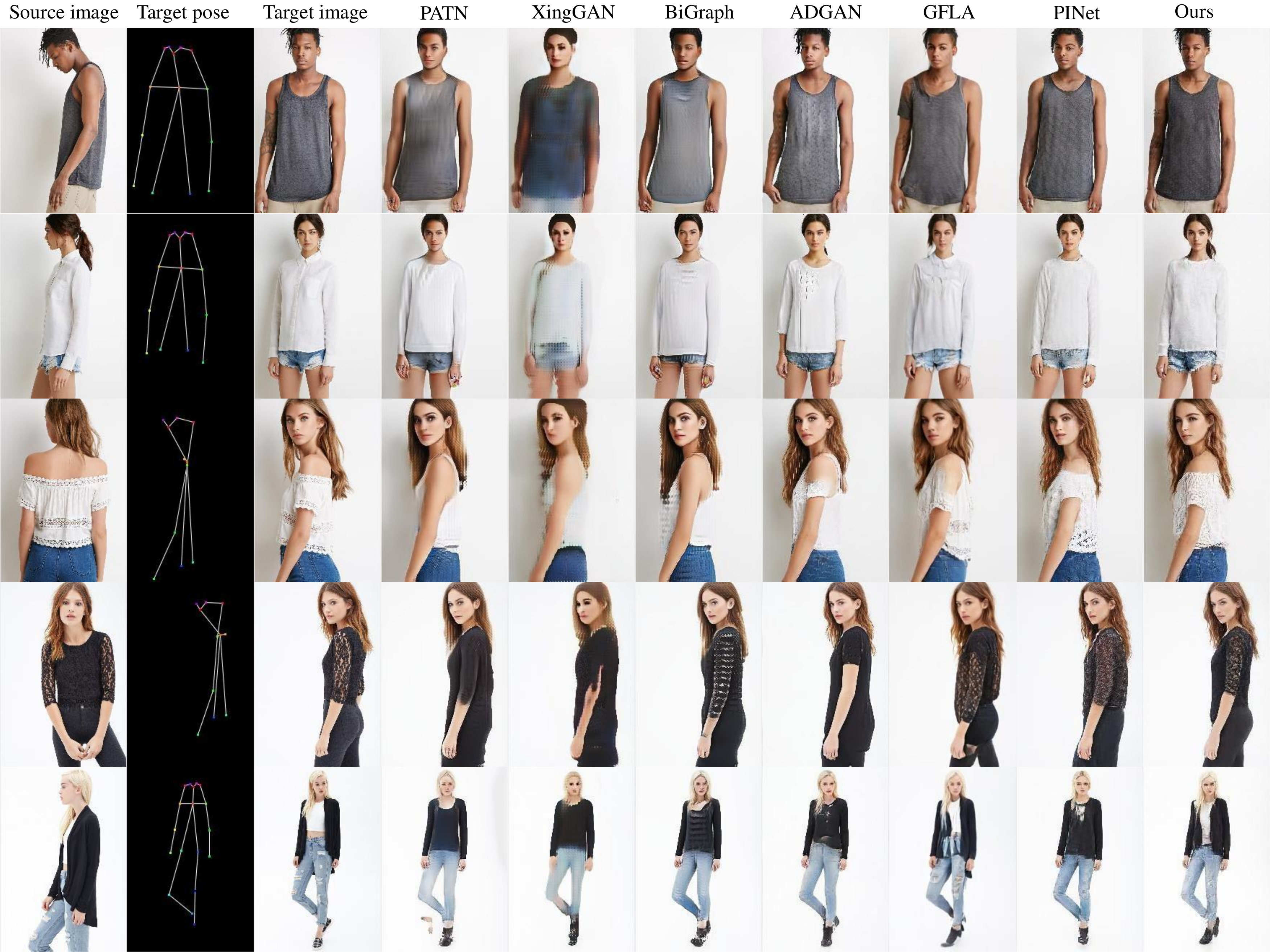}
		\end{center}
		\caption{Qualitative comparisons with state-of-the-art methods. From left to right are the results of PATN \cite{patn}, XingGAN \cite{xinggan}, BiGraph \cite{bi}, ADGAN \cite{adgan}, GFLA \cite{gfla}, PInet \cite{pinet} and ours, respectively. }
		\label{fig_com}
		\vspace{-0.2cm}
	\end{figure*}
	
	\subsection{Loss Functions}
	We first train the parsing generator and the image generator respectively and then end-to-end train our full model.
	In the following, we will describe the loss function of two generators in details.
	
	\noindent
	\textbf{Parsing Generator Loss.}
	The full loss function of the parsing generator can be formulated as:
	\begin{equation}
	\mathcal{L}_{parsing} =  \mathcal{L}_{cross} + \lambda_{p\ell} \mathcal{L}_{\ell_1}.
	\vspace{-0.2cm}
	\end{equation}
	where  $\lambda_{p\ell}$ is the coefficient of the $\ell_1$ item.
	
	In order to generate reasonable human parsing maps we apply  $\ell_1$ distance loss between the generated parsing map and the target parsing map:
	\begin{equation}
	\mathcal{L}_{\ell_1}=||S_g-S_t||_1.
	\vspace{-0.2cm}
	\end{equation} 
	
	Besides, to generate high-quality human parsing maps, we also use cross-entropy loss $\mathcal{L}_{cross}$, which is defined as:
	
	\begin{equation}
	\mathcal{L}_{cross} = - \frac{1}{N} \sum_{i=0}^{N-1}  S_{t_i}  \log (\textrm{Softmax}(S_{g_i})),
	\vspace{-0.2cm}
	\end{equation}
	where $N$ is the number of categories of labels ($N=8$ in our case).

	\noindent
	\textbf{Image Generator Loss.}
	The full loss used for training the image generator consists of correspondence loss, reconstruction $\ell_1$ loss, perceptual loss, style loss, and adversarial loss, defined as:
	\begin{equation}
	\mathcal{L}_{image} = \lambda_c \mathcal{L}_{cor} +\lambda_\ell \mathcal{L}_{\ell} + \lambda_p \mathcal{L}_{per} + \lambda_s \mathcal{L}_{style} + \lambda_a \mathcal{L}_{\textrm{adv}}, 
	\vspace{-0.2cm}
	\end{equation}
	where $\lambda_c$, $\lambda_\ell$, $\lambda_p$, $\lambda_s$ and $\lambda_a$ are weights that balance contributions of individual loss terms.
	
	The correspondence loss is used to constrain the generated feature $F_n$ and the pre-trained VGG-19 \cite{vgg} feature of the target image to align in the same domain, which is defined as:
	\begin{equation}
	\mathcal{L}_{cor} = ||F_n- {\phi_i}(I_t)||_2.
	\vspace{-0.2cm}
	\end{equation} 
	In practice, we use the feature from $conv3\_1$ of VGG-19 to compute the correspondence loss.
	
	The reconstruction $\ell_1$ loss is used to encourage the generated image $I_g$ to be similar with the target image $I_t$ at the pixel level, which is defined as:
	\begin{equation}
	\mathcal{L}_{\ell_1} = ||I_g - I_t||_1.
	\vspace{-0.2cm}
	\end{equation}
	We also adopt perceptual loss and style loss \cite{perceptual} to generate more realistic images. The perceptual loss calculates the $\ell_1$ distance between activation maps of the pre-trained VGG-19 network, which can be written as:
	\begin{equation}
	\mathcal{L}_{per} = \sum_{i} ||\phi_i(I_g) - \phi_i(I_t)||_1.
	\vspace{-0.2cm}
	\end{equation}
	We adopt the feature of [$relu1\_1$, $relu2\_1$, $relu3\_1$, $relu4\_1$, $relu5\_1$] with the same weight.
	The style loss measures the statistical difference of the activation maps between the generated image $I_g$ and the target image $I_t$, which is formulated as:
	\begin{equation}
	\mathcal{L}_{style} = \sum_{j} ||G_j^\phi(I_t)-G_j^\phi(I_g)||_1.
	\vspace{-0.2cm}
	\end{equation}
	In practice, we use the feature of [$relu2\_2$, $relu3\_4$, $relu4\_4$, $relu5\_2$] with the same weight.
	The adversarial loss with discriminator $D$ is employed to penalize the distribution difference between generated (fake) images $I_g$ and target (real) images $I_t$, which is written as:
	\begin{equation}
	\mathcal{L}_{adv} = \mathbb{E}[\log(1-D(I_g))]+\mathbb{E}[\log D(I_t)].
	\vspace{-0.2cm}
	\end{equation}

	\vspace{-0.2cm}
	\section{Experimental Results}
	\vspace{-0.1cm}
	
	\noindent
	\textbf{Dataset.} We conduct our experiment on DeepFashion In-shop Clothes Retrieval Benchmark \cite{deepfashion}, which contains 52712 images with the resolution of 256 $\times$ 256. 
	The images vary in terms of poses and appearances.
	We split the data  with the same configuration as PATN \cite{patn} and collect 101,966 pairs of
	images for training and 8,750 pairs for testing. 
	Note that the person identities in the test set are different from those in the training set.
	
	\noindent
	\textbf{Metrics.} We use Learned Perceptual Image Patch Similarity (LPIPS) \cite{lpips} to measure the distance between the generated image and the ground-truth image in the perceptual level.
	Meanwhile, Peak Signal to Noise Ratio (PSNR) is employed to compute the error between the generated image and the ground-truth image in the pixel level.
	Besides, we adopt Fr{$\acute{\textup{e}}$}chet Inception Distance (FID) \cite{fid} to compute the distance between distributions of the generated images and the ground-truth images, which is used to measure the realism of the generated images. 
	
	\subsection{Person Image Synthesis in Different Poses}
	Human pose transfer aims at synthesizing new images for the same source person in different poses.
	Figure \ref{fig_arb} shows some visual results generated by our method. 
	Given a source image and some different poses extracted from images in the test set, our model generates realistic images with fine details. 
	
	\begin{table}[!t]
		\renewcommand{\arraystretch}{1.3}
		\small%
		\setlength{\tabcolsep}{2.5mm}
		\begin{center}
			\caption{Quantitative comparison with state-of-the-art methods$^1$.}\label{tab:tab_com}
			\begin{tabular}{|c|c|c|c|}
				\hline
				{Model} & FID $\downarrow$ & LPIPS $\downarrow$ & PSNR $\uparrow$ \\
				
				\hline
				PATN \cite{patn} & 23.70 & 0.2520 & 31.16 \\
				BiGraph \cite{bi} & 24.36 & 0.2428 & \textbf{31.38} \\
				XingGAN \cite{xinggan} & 41.79 & 0.2914 & 31.08 \\
				GFLA \cite{gfla} & \underline{14.52} & 0.2219 & 31.28 \\
				ADGAN \cite{adgan} & 16.00 & 0.2242 & 31.30\\
				PINet \cite{pinet} & 15.28 & \underline{0.2152} & \underline{31.31}\\
				\hline
				Ours & \textbf{13.61} & \textbf{0.2059} & \textbf{31.38}  \\

				\hline
			\end{tabular}
			
		\end{center}
		\vspace{-0.2cm}
	\end{table}
	\footnotetext[1]{Note that we take the images resized from 256 $\times$ 176 to 256 $\times$ 256 as the inputs of GFLA.}

	\begin{table}[!t]
		\renewcommand{\arraystretch}{1.3}
		\small%
		\setlength{\tabcolsep}{2.5mm}
		\begin{center}
			\caption{Quantitative results of ablation study.}\label{tab:tab_aba}
			\begin{tabular}{|c|c|c|c|}
				\hline
				{Model} & FID $\downarrow$ & LPIPS $\downarrow$ & PSNR $\uparrow$ \\
				
				\hline
				Global-Enc & 15.21 & 0.2137 & 31.35 \\
				Local-Enc  & 15.50 & 0.2138 & 31.30 \\
				w/o SN  & \underline{14.15} & \underline{0.2071} & \textbf{31.43}\\
				\hline
				Full & \textbf{13.61} & \textbf{0.2059} & \underline{31.38}  \\

				\hline
			\end{tabular}
		\end{center}
		\vspace{-0.6cm}
	\end{table}

	\subsection{Comparison with State-of-the-art Methods}
	
	We conduct qualitative and quantitative comparisons with several state-of-the-art methods.
	
	\noindent
	\textbf{Qualitative Comparison.} We compare the visual results of our method with six state-of-the-art methods: PATN \cite{patn}, XingGAN \cite{xinggan}, BiGraph \cite{bi}, ADGAN \cite{adgan}, GFLA \cite{gfla} and PINet \cite{pinet}. 
	Figure \ref{fig_com} shows some examples.
	PATN, XingGAN, and BiGraph fail to generate sharp images and cannot keep the consistency of shape and texture due to the use of sparse correspondence extracted from keypoints.
	The flow-based method, GFLA, preserves the detailed texture in the source image.
	However, it is difficult to obtain reasonable results for the invisible regions of the source image.
	ADGAN and PINet succeed in keeping shape consistency and predicting reasonable shapes of clothing, but they lose spatial context relationships.
	Our model uses spatial-aware normalization to retain spatial context relationships, and hence can generate more natural and sharper images (the second and fourth rows).
	Besides, our model retains the shape (the first and third rows)  and predicts more reasonable results (the second and fifth rows).
	
	\noindent 
	\textbf{Quantitative Comparison.} Table \ref{tab:tab_com} gives the quantitative results on the 8750 test images compared with six state-of-the-art methods: PATN \cite{patn}, XingGAN \cite{xinggan}, BiGraph \cite{bi}, ADGAN \cite{adgan}, GFLA \cite{gfla} and PINet \cite{pinet}. 
	Our results get the best PSNR score, which measures the error in pixel level. 
	Besides, we introduce LPIPS to compute the similarity  in perceptual level and FID to measure the realism of the generated images. 
	Our results achieve the best performance in terms of both FID and LPIPS, which indicates that our model not only generates more realistic images but also keeps the best consistency of shape and texture. 
	
	\begin{figure}[!t]
		
		\centering
		\includegraphics[scale=0.45]{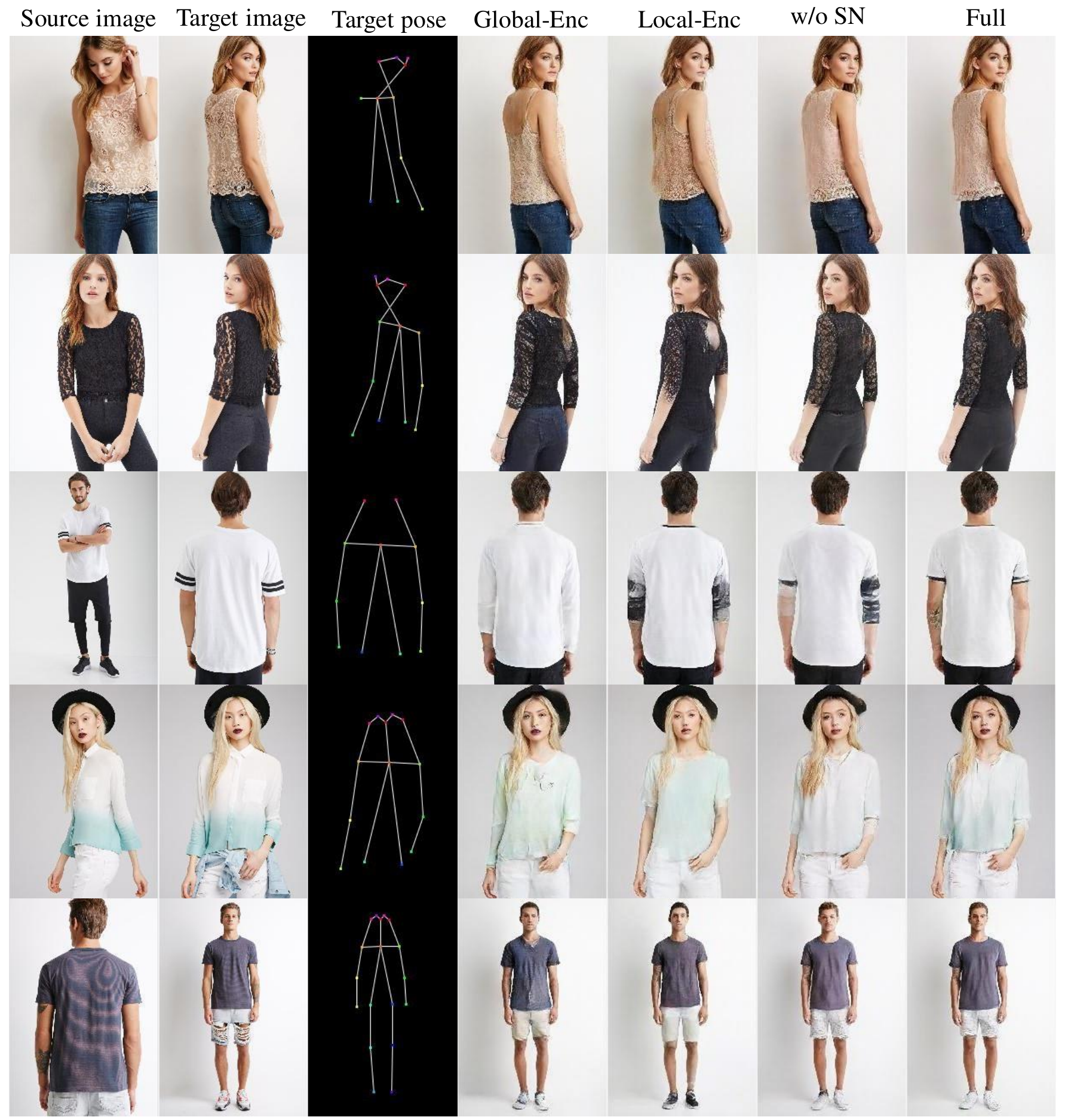}
		\caption{Qualitative results of ablation study. }
		\label{fig_aba}
		\vspace{-0.5cm}
	\end{figure}
	
	\begin{figure*}[!ht]
		
		\centering
		{
			\begin{minipage}[t]{0.5\linewidth}
				\centering
				\label{fig_tta}
				\includegraphics[scale=0.25]{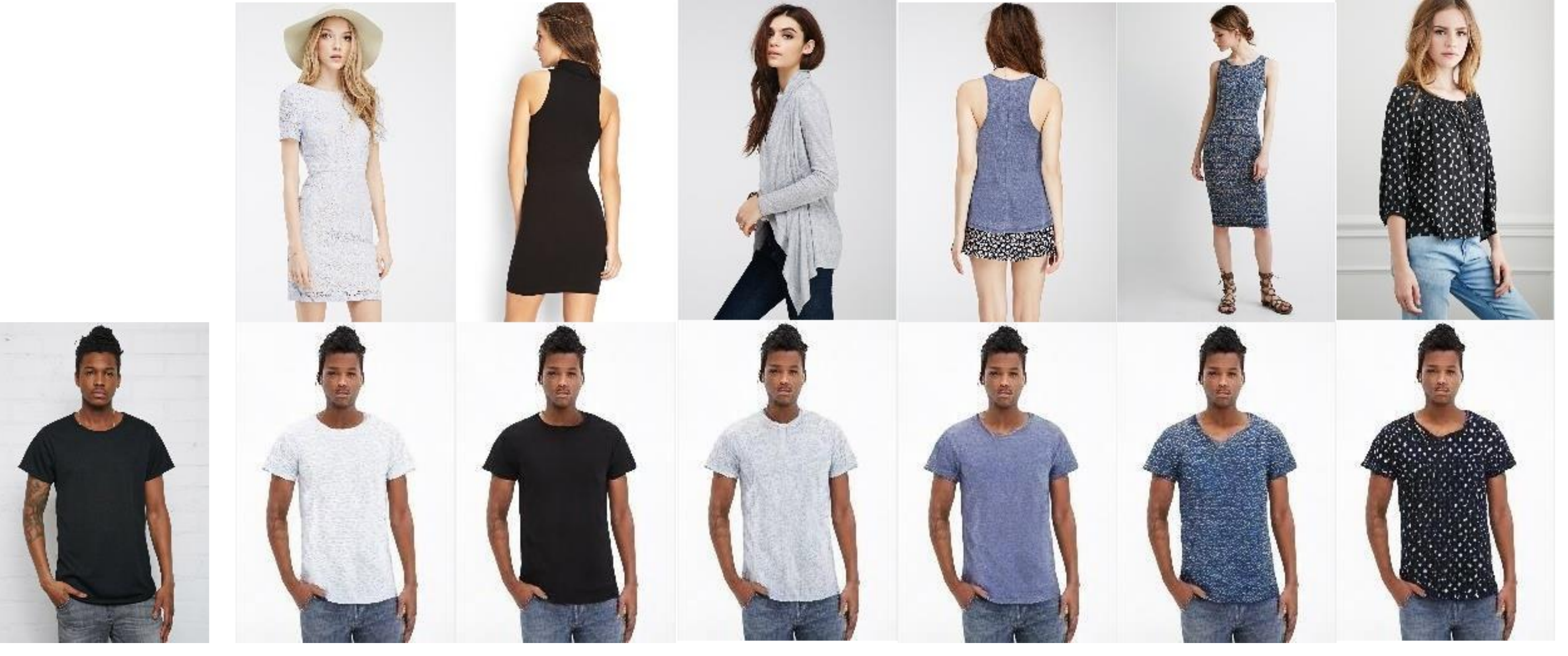}
			\end{minipage}%
			\begin{minipage}[t]{0.5\linewidth}
				\centering
				\label{fig_ttb}
				\includegraphics[scale=0.25]{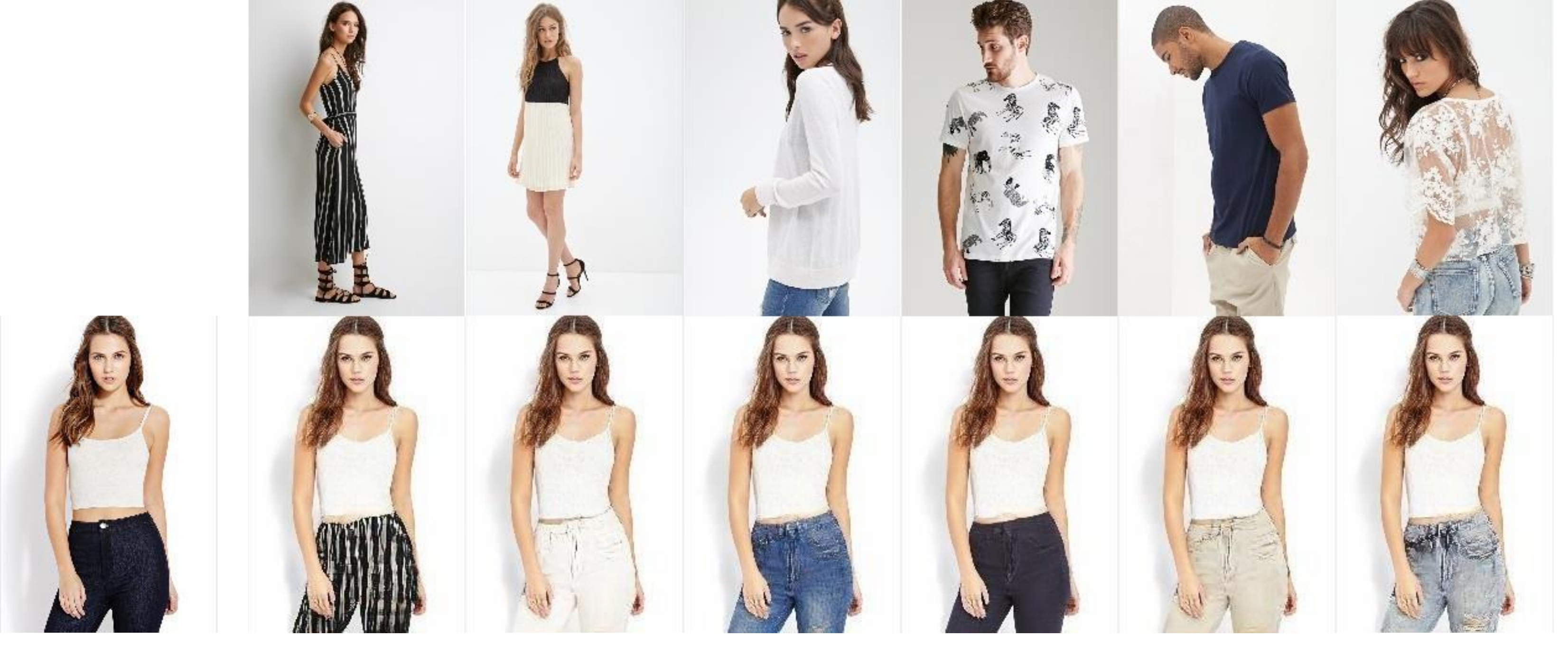}
			\end{minipage}%
			
		}%
		
		\caption{Texture transfer results. The left figure and the right figure show the results of transfer the texture of upper clothes and pants, respectively. In each figure, the first row shows the reference images, and the second row shows the results generated by our method. The first column of each figure shows the source image.}
		\label{fig_tt}
		\vspace{-0.2cm}
	\end{figure*}
	
	\subsection{Ablation Study}
	We train several ablation models to prove our hypotheses and validate the effect of our improvements.
	
	\noindent
	\textbf{Global Encoding Model (Global-Enc).} The global encoding model 
	replaces per-region styles with the same global feature extracted from the source image using the global average pooling. 
	
	\noindent
	\textbf{Local Encoding Model (Local-Enc).} The local encoding model adopts per-region encoding. However, for the style of invisible regions, the style code is set to be zero.
	
	\noindent
	\textbf{The Model without Spatial-aware Normalization (w/o SN).} This model is designed to measure the contribution of spatial-aware normalization described in Section \ref{sec_sn}. We train this model in the same configuration as our full model.
	
	\noindent
	\textbf{Full Model (Full).} We use joint global and local per-region encoding and spatial-aware normalization in this model.
	
	
	Quantitative results on the 8750 test images are shown in Table \ref{tab:tab_aba}.
	As shown in  Table \ref{tab:tab_aba}, compared with the global encoding model and local encoding model, the joint global and local per-region encoding and normalization improve the performance by predicting the style of each region, especially for the authenticity of generated images and the similarity to the ground-truth images.
	Besides, due to the detailed images generated by our full model, our full model has a slightly lower PSNR than the model without spatial-aware normalization  in the pixel level, but it gains the best performance in similarity in the perceptual level and generates the most realistic images.
	Figure \ref{fig_aba} shows some visual results of ablation study. 
	With joint global and local encoding and normalization, our method can maintain the exact style of visible regions (in the first and second rows) and predict reasonable styles of invisible regions (natural face and shoes in the fifth row).
	With spatial-aware normalization, our model generates sharper images with the spatial context relationship (in the second, third and fourth rows).

	\vspace{-0.2cm}
	\section{Applications}
	
	Benefiting from our two-stage framework and per-region style encoding, our model decouples the shape and style of clothing and can be applied to various person image editing applications. 
	
	\begin{figure}[!t]
		
		\centering
		\includegraphics[scale=0.28]{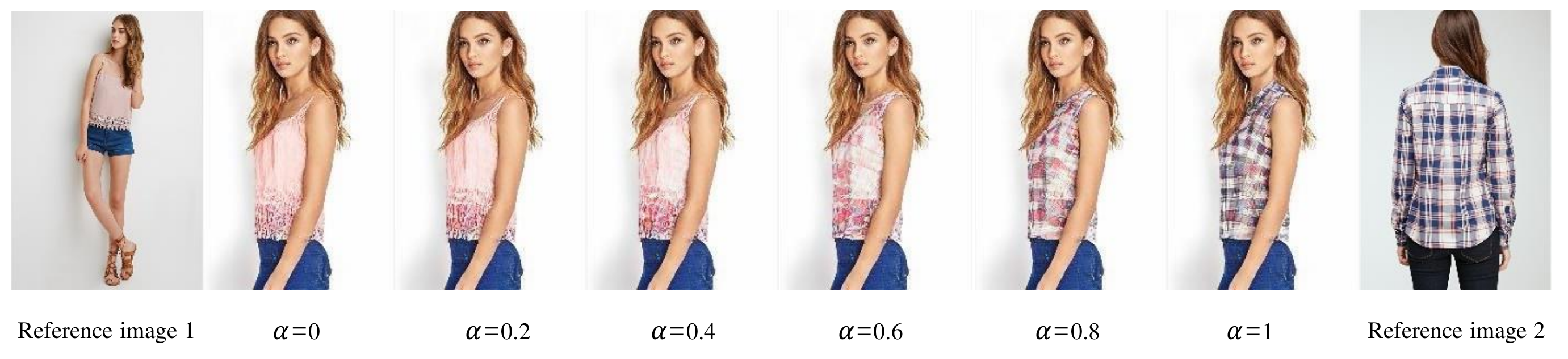}
		\caption{Texture interpolation results.}
		\label{fig_inter}
		\vspace{-0.3cm}
	\end{figure}
	
	\begin{figure}[!t]
		
		\centering
		\includegraphics[scale=0.26]{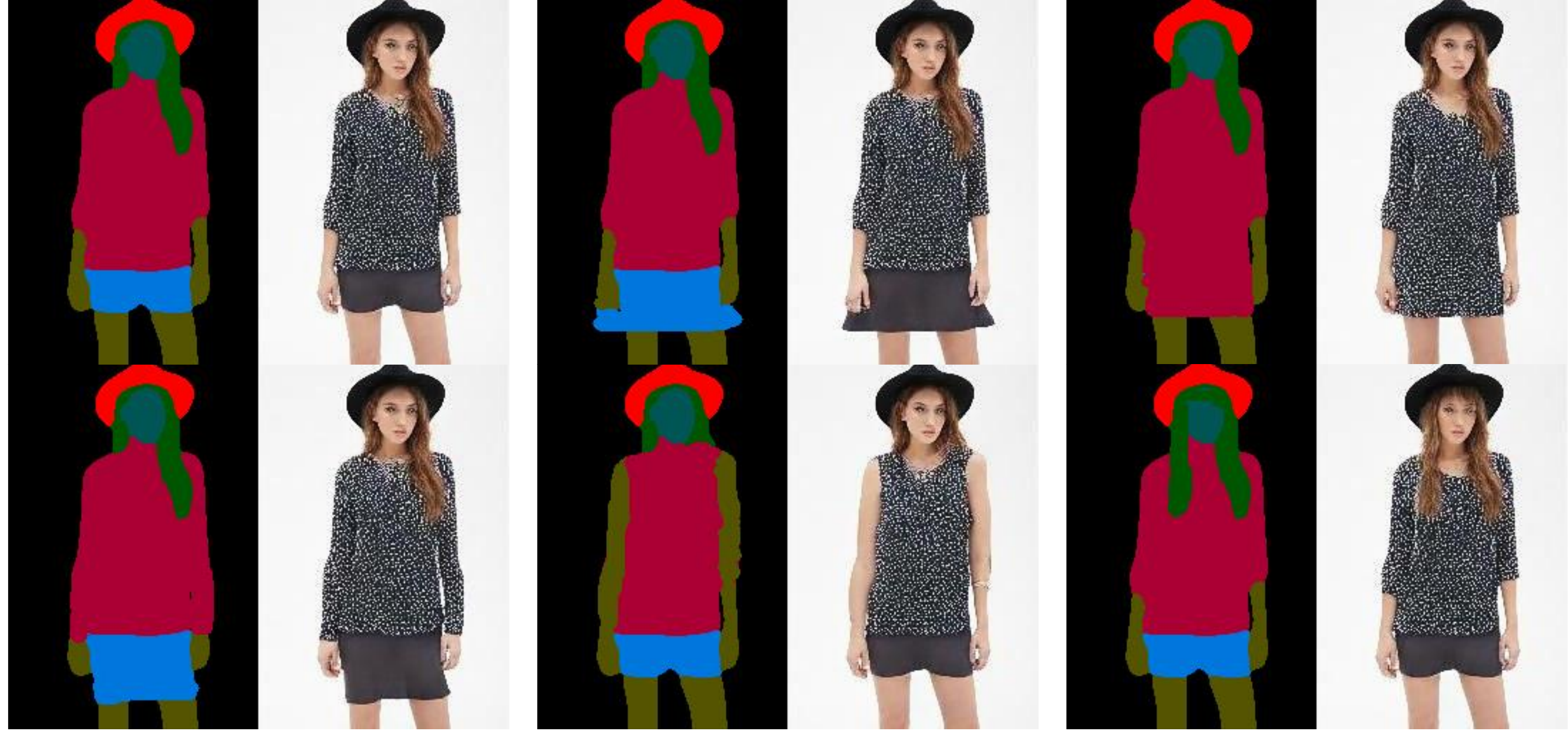}
		\caption{Region editing results. }
		\label{fig_edit}
		\vspace{-0.3cm}
	\end{figure}

	\noindent
	\textbf{Texture Transfer}
	The per-region style of our generated images is controllable by the style code $P(S_{s_j})$. 
	Therefore, we can change the style of each region by replacing the style code of the corresponding region in  $P(S_{s_j})$.
	Specifically, given a reference image, we can extract its style code and transfer the per-region style to our generated image.
	Figure \ref{fig_tt} shows some visual results of texture transfer. 
	We transfer the texture of upper clothes or pants of the reference images to our generated images.
	Our model generates natural images with detailed texture.

	\noindent
	\textbf{Texture Interpolation}
	We control the per-region styles  of our generated images by latent code, which can move along the manifold of textures from one style to another style. 
	Taking texture transfer of upper clothes as an example, we interpolate the styles of upper clothes ($j$-th in the number of categories) from one image $I_{s_1}$ to another image $I_{s_2}$. The style $t$ of the upper clothes of the generated image is defined through linear blending as:
	\begin{equation}
	t = (1-\alpha)  P(S_{s_{1j}}) + \alpha  P(S_{s_{2j}}).
	\end{equation}
	As shown in Figure \ref{fig_inter}, the texture of the upper clothes gradually changes from the style of the left reference image to that of the right reference image.

	\noindent
	\textbf{Region Editing}
	Since we use a human parsing map as the intermediate result, we can edit the generated images by editing the input parsing of the image generator. 
	Specifically, given a source image and its parsing map, we can flexibly edit the parsing map to automatically synthesize the person image as we desired. 
	As shown in Figure \ref{fig_edit}, we can change the style of clothing (\emph{e.g.}, T-shirt to waistcoat, pants to dress, and long hair to short hair).
	Our model generates natural images consistent with the edited parsing map.


	\vspace{-0.2cm}
	\section{Conclusion}
	
	In this paper, we propose PISE, a novel two-stage generative model for person image synthesis and editing.
	Our method first synthesizes a human parsing map and then generates the target image by joint global and local encoding and normalization  and spatial-aware normalization.
	Experimental results demonstrate that our model achieves promising results with detailed texture and consistent shape of clothing.
	Besides, the ablation study also verifies the effectiveness of each designed component.
	Our model can also be applied in various applications such as texture transfer and region editing, and generates natural images. 
	In the future, we will try to generalize our framework to deal with video generation conditioned on a reference image.

	\noindent
	\textbf{Acknowledgments. }This work was supported in part by the National Natural Science Foundation of China (61771339) and Tianjin Research Program of Application Foundation and Advanced Technology (18JCYBJC19200).
	
	{\small
		\bibliographystyle{ieee_fullname}
		\bibliography{egbib}
	}
	
\end{document}